\documentclass[runningheads]{llncs}
\usepackage{graphicx}

\usepackage{amsmath,amssymb,amsfonts}
\usepackage{algorithmic}
\usepackage{graphicx}
\usepackage{textcomp}
\usepackage{xcolor}
\usepackage{enumitem}
\usepackage{booktabs} % for much better looking tables
\usepackage{url} % Useful for inserting web links nicely
\usepackage[bookmarks,unicode,hidelinks]{hyperref}
\usepackage{breakurl}
\usepackage{color}
\definecolor{light}{rgb}{0.5, 0.5, 0.5}
\def\light#1{{\color{light}#1}}

\def\BibTeX{{\rm B\kern-.05em{\sc i\kern-.025em b}\kern-.08em
    T\kern-.1667em\lower.7ex\hbox{E}\kern-.125emX}}

\makeatletter
\def\thickhline{%
  \noalign{\ifnum0=`}\fi\hrule \@height \thickarrayrulewidth \futurelet
   \reserved@a\@xthickhline}
\def\@xthickhline{\ifx\reserved@a\thickhline
               \vskip\doublerulesep
               \vskip-\thickarrayrulewidth
             \fi
      \ifnum0=`{\fi}}
\makeatother

\newlength{\thickarrayrulewidth}
\begin{document}
\title{Enhancing Romanian Offensive Language Detection through Knowledge Distillation, Multi-Task Learning, and Data Augmentation}
\titlerunning{Enhancing Romanian Offensive Language Detection}

\author{Vlad-Cristian Matei \and
Iulian-Marius Tăiatu \and
Răzvan-Alexandru Smădu \and
Dumitru-Clementin Cercel\textsuperscript{\rm *}}

\authorrunning{V.-C. Matei et al.}

\institute{
Faculty of Automatic Control and Computers, National University of Science and Technology POLITEHNICA Bucharest, Romania
}

\def\thefootnote{*}\footnotetext{Corresponding author: dumitru.cercel@upb.ro.}
\def\thefootnote{\arabic{footnote}}

% \and
% \email{lncs@spr
% inger.com}\\
\maketitle              % typeset the header of the contribution
\begin{abstract}
This paper highlights the significance of natural language processing (NLP) within artificial intelligence, underscoring its pivotal role in comprehending and modeling human language. Recent advancements in NLP, particularly in conversational bots, have garnered substantial attention and adoption among developers. This paper explores advanced methodologies for attaining smaller and more efficient NLP models. Specifically, we employ three key approaches: (1) training a Transformer-based neural network to detect offensive language, (2) employing data augmentation and knowledge distillation techniques to increase performance, and (3) incorporating multi-task learning with knowledge distillation and teacher annealing using diverse datasets to enhance efficiency. The culmination of these methods has yielded demonstrably improved outcomes.

\keywords{Offensive Language \and Knowledge Distillation \and Multi-Task Learning \and Data Augmentation}
\end{abstract}
\section{Introduction}
In recent years, there has been a remarkable rise in the prominence of natural language processing, primarily attributable to the success of conversational bots like ChatGPT\footnote{\url{https://www.openai.com/chatgpt}}. These models have achieved impressive performance through extensive training on massive datasets. However, the battle against fake news, offensive language, and abusive content on social networks remains challenging due to the overwhelming volume of user-generated data, necessitating the development of automatic detection systems \cite{eu_law}. Moreover, the intricate nature of identifying offensive language stems from the nuanced considerations of contextual factors, multiple meanings, and emerging expressions \cite{abuse_survey}. Further advancements in this field are still required to tackle these challenges effectively.

Exploiting offensive language has garnered significant global attention, with trained models demonstrating notable performance achievements \cite{abuse_same}. However, when considering the specific context of the Romanian language analysis, the existing datasets are constrained in size and availability \cite{rofb}. To understand crucial and intricate characteristics comprehensively, models require a diverse and well-balanced collection of qualitative examples. Additionally, the architectures underlying these models comprise millions of parameters, resulting in substantial computational and resource requirements \cite{modelcompr2}. Consequently, although automatic detection of offensive language remains relevant, challenges arise in adapting these models for mobile devices or embedded systems due to limitations in speed, space, and resource constraints \cite{arhitectura_bert}.

The primary objective of this study is to develop an automatic offensive language detection model encompassing three distinct offensive categories (i.e., \textit{Insult}, \textit{Profanity}, and \textit{Abuse}) and a neutral class \textit{Other}. Initially, we employ the knowledge distillation (KD) method \cite{kd} to transfer information \cite{transferlr} from a high-parameter model to a more compact architecture \cite{modelcompr2}. Subsequently, the obtained results undergo meticulous analysis, complemented by an exploration of various data augmentation strategies: generative text \cite{gpt} by employing RoGPT-2 \cite{rogpt}, ASDA \cite{asda}, MixUp \cite{mixup}, and noisy student \cite{noisy_student}. These techniques enhance the model's performance by introducing nuanced variations or controlled noise injection.

In addition, this study prioritizes performance optimization rather than model size reduction. Accordingly, knowledge distillation is employed on an architecture with an equivalent number of parameters \cite{arhitectura_bert}. Furthermore, a multi-task learning (MTL) approach \cite{mtl,mtl2,caragea} integrates information from three auxiliary tasks associated with sentiment analysis \cite{laroseda}, emotions analysis \cite{redv2}, and sexist language \cite{coroseof}. We comprehensively evaluate the effectiveness of this architecture in efficiently assimilating and integrating the acquired information. Furthermore, our study assesses how these auxiliary datasets contribute to a more comprehensive understanding of the Romanian language, particularly in the context of the initial problem. To summarize, the contributions of this work are: (i) evaluating the knowledge distillation method to obtain a more compact and faster model and showing that utilizing diverse data augmentation techniques improves performance, and (ii) performance enhancement by applying multi-task learning with knowledge distillation and teacher annealing \cite{tannealing}, integrating information from three additional datasets.

\section{Related Work}

Automatic offensive language detection poses a challenge of global interest \cite{eu_law,abuse_survey}, with attempts to address the issue using various means, both classic machine learning methods and deep learning approaches \cite{zampieri-etal-2019-semeval}. Offensive language detection was proposed at several workshops, including SemEval-2019 \cite{zampieri-etal-2019-semeval} and GermEval-2019 \cite{abuse_same}. Baseline models such as support vector machines have been evaluated on offensive and sexist tweets \cite{coroseof}. In addition, \cite{emohs} explored neural network approaches such as long short-term memory and convolutional neural networks on hate speech, showing there is room for improvement in these systems.

To optimize recent models, transfer learning techniques \cite{transferlr} have been proposed, including multi-task learning \cite{mtl,mtl2} and knowledge distillation \cite{kd,modelcompr2}, with applications extending to the domain of offensive language \cite{vlad2019sentence}. AngryBERT \cite{angryb} combined MTL with the Bidirectional Encoder Representations from Transformer (BERT) model \cite{bert} to jointly learn hate speech detection along with emotion classification and target identification as secondary tasks, enhancing overall performance. In \cite{mtla}, the authors investigate bridging differences in annotation and data collection of hate speech and abusive language in tweets, including various annotation schemes, labels, and geographic and cultural influences. To harness the benefits of both methods, models combining multi-task learning with knowledge distillation \cite{mtkd,mtkd_born,carageamtkd,caragea2} or teacher annealing have been proposed \cite{tannealing}. This framework was also employed alongside the teacher annealing option, albeit for empathy detection \cite{caragea}.
In contrast to other works, we combine all these methods to address the automatic detection of offensive language, particularly in the Romanian language, and compare them individually and through an ablation study.

\section{Method}

\subsection{Fine-tuning BERT}

In this work, we base our models on the BERT architecture \cite{bert}. The baseline involves fine-tuning BERT for automatic offensive language detection. It adjusts the weights obtained from a pre-trained BERT to identify and classify different subtypes of offensive language accurately.

The initial step establishes a manually annotated dataset $\mathcal{D} = {\big\{ (x_{i}, y_{i}) \big\}}_{i=1:N}$ with $N$ examples, namely RO-Offense\footnote{\url{https://huggingface.co/datasets/readerbench/ro-offense}}, that assigns labels $y_{i}$ (i.e., \textit{Insult}, \textit{Abuse}, \textit{Profanity}, and \textit{Other}) to each text comment $x_{i}$. These labels enable a more nuanced understanding of the offensive language, moving beyond a simplistic binary classification.
To accommodate the classification task with $K=4$ possible classes, we add a fully connected layer on top of the last layer of BERT's architecture. This fully connected layer consists of $K$ neurons, each corresponding to one of the predicted classes. Then, a softmax layer computes the probability distribution $p_i$ over predicted classes for the given input text $x_i$.

During the training process, the weights are updated to minimize the prediction error by employing the cross-entropy loss function $\mathcal{L}_{CE}$. This loss quantifies the dissimilarities between the predicted and true labels as follows:
\begin{equation}
    \mathcal{L}_{CE} = -\frac{1}{N} \sum_{i=1}^{N} \sum_{k=1}^{K} y_{i,k} \log{(p_{i,k})}
\end{equation}
where $y_{i,k}$ is the kth class label from the one-hot encoding for the ground truth $y_i$, and $p_{i,k}$ is the kth class probability from the model's softmax output $p_i$.

\subsection{Data Augmentation}
Data augmentation is a technique employed to enrich the diversity of features in the dataset without the need for additional data collection \cite{dataug}. Its objective is to introduce changes in input data to enhance the models' capacity for generalization. This approach offers several benefits, including improved model performance, acting as a regularization technique, enhancing model robustness, and addressing class imbalance \cite{guo_mixup,asda,dataug}. Our work employs several data augmentation techniques.
%However, it increases computational and storage requirements, while inappropriate augmentation can lead to poor models.

\textbf{RoGPT-2.} Generative Pre-trained Transformer (GPT) models \cite{gpt} leverage the linguistic knowledge and comprehensive understanding of language structures to produce varied texts that encompass the intricacies and diversity observed in real-world texts. While existing methods like Easy Data Augmentation \cite{eda} focus on simple transformations applied to the existing text, GPT models offer the advantage of generating creative and meaningful texts, enhancing the model's robustness through variety. However, the generated texts may significantly alter the original meaning. To mitigate this risk, we employ the RoGPT-2 \cite{rogpt} model. RoGPT-2 is the Romanian version of the GPT-2 model \cite{radford2019language}, pre-trained on a large 17 GB Romanian text dataset. We provide 70\% of the context to control the level of text augmentation, allowing the model to generate a sequel and encouraging controlled and coherent text generation.

\textbf{ASDA.} Auxiliary Sentence-based Data Augmentation (ASDA) \cite{asda} utilizes conditional masked language modeling \cite{cmlm} to generate augmented examples. First, it works by selecting an example \textit{E1} from the training dataset and then choosing another example \textit{E2} with the same class \textit{[LABEL]} as \textit{E1}. Next, we construct the context using the following template: ``\texttt{The next two sentences are \textit{[LABEL]}.} \texttt{The first sentence is: \textit{E1}.} \texttt{The second sentence is: \textit{E2}.}". We apply random masking of a set of words within the final sentence, E2. The masking process occurs with a predefined probability and depends on the sentence length.

\textbf{MixUp.} MixUp \cite{mixup} is a data augmentation method intended to boost diversity while lowering the possibility of generating incorrect examples. It creates a more robust generalization by linearly interpolating between various dataset examples. The technique randomly selects two examples (i.e., $(x^i, y^i)$ and $(x^j, y^j)$) from the dataset, where $x^i$ and $x^j$ represent the encoded inputs, whereas $y^i$ and $y^j$ are the one-hot encoded labels. New examples $(\hat{x}, \hat{y})$ are generated using the following formulas:

\begin{equation}
\hat{x} = \lambda x_i + (1 - \lambda)x_j,
\end{equation}
\begin{equation}
\hat{y} = \lambda y_i + (1 - \lambda)y_j,
\end{equation}
where $\lambda \in [0, 1]$ is a hyperparameter that controls the interpolation process.
In this work, we employ the MixUp technique at two distinct levels\footnote{\url{https://github.com/xashru/mixup-text}} \cite{guo_mixup}:

\begin{itemize}
\item{\textbf{MixUp Encoder}: Interpolates the representations of the two input examples before passing them through the classification layer.}
\item{\textbf{MixUp Sentence}: Interpolates the representations of the two inputs after the classification layer but before the softmax activation function.}
\end{itemize}

\textbf{Noisy Student.} Noisy student \cite{noisy_student} is employed in noisy student training \cite{noisy_training}, where a teacher model generates pseudo-labels for unlabeled data, and a student model is trained on these pseudo-labels. The objective is to introduce natural noise and enhance the robustness of the model. In this research, we apply two non-aggressive methods \cite{noisy_student}:

\begin{itemize}
\item{\textbf{Word Drop}: We choose, with a probability $\alpha$, that every word in a sentence has a fixed chance of 30\% to be removed. It is guaranteed that at least one word will be deleted, but no more than ten words in total.}
\item{\textbf{Sentence Drop}: In cases where the example contains at least two sentences, we remove one sentence from the text with the same probability $\alpha$.}
\end{itemize}

\subsection{Multi-Task Learning Model} \label{CC}

MTL \cite{mtl2,mtl} is a method that learns several related tasks simultaneously within a single framework to enhance target task performance.  The fundamental concept behind MTL is based on the observation that learning similar tasks in parallel can facilitate quicker adaptation, leveraging common principle knowledge \cite{mtl}. Building upon this intuition, MTL learns shared representations that encapsulate the underlying essence of the information while capturing the specific characteristics of each task up to a level beneficial for the main task \cite{mtl2}. This approach can be viewed as a subcategory of TL, as both leverage the knowledge from related tasks \cite{transferlr}. However, MTL offers several advantages, including leveraging information learned from different tasks and transferring knowledge across diverse datasets \cite{mtl2}. It also helps mitigate overfitting by regularizing the network and preventing single tasks from dominating the learning process \cite{mtl2}.

Inspired by \cite{mtl2,caragea}, our MTL architecture consists of shared lower layers derived from BERT common to all tasks and task-specific layers added to this backbone. A softmax activation function follows each task-specific layer to obtain the probability distribution for the corresponding task. In this work, the primary task of offensive language detection is supported by three auxiliary tasks, namely emotion classification, sentiment analysis, and sexist language detection. These additional tasks are considered to be correlated with the target domain, as previous research \cite{abuse_survey,emohs,emohs2} has demonstrated their relevance to offensive language detection. Therefore, let $\mathcal{D}^\tau = {\big\{ (x_{i}^{\tau}, y_{i}^{\tau}) \big\}}$ be the training dataset for a task $\tau$. The multi-task loss $\mathcal{L}_{MTL}$, calculated for a model $\theta$, is defined as follows \cite{caragea}:

\begin{equation}
\mathcal{L}_{MTL}(\theta) = \sum_{\tau=1}^{4} \sum_{(x_{i}^{\tau}, y_{i}^{\tau}) \in \mathcal{D}^\tau} \ell (y_{i}^\tau, f^{\tau}(x_{i}^{\tau}; \theta))
\end{equation}
where $\ell$ is either the binary cross-entropy or the cross-entropy loss depending on the task, and $f^{\tau}(x_{i}^{\tau}, \theta)$ denotes the output of the model $\theta$ for the task $\tau$. We employ cross-entropy loss $\mathcal{L}_{CE}$ for offensive language detection, sentiment classification, and sexist language detection. For the emotion analysis dataset, which comprises seven classes, we use the one-hot encoding representation for the labels, and thus, the loss function $\ell$ is the binary cross-entropy loss defined as:

\begin{equation}
\mathcal{L}_{BCE} = -\frac{1}{NK} \sum_{i=1}^N \sum_{k=1}^K \left[ y_{i,k} \cdot \log (p_{i,k})
+ (1 - y_{i,k}) \cdot \log (1 - p_{i,k}) \right]
\end{equation}
where $y_{i,k}$ denotes the kth class label from the one-hot ground truth label, and $p_{i,k}$ is the model's prediction for the kth class.

\subsection{Knowledge Distillation Models} \label{KD}

\textbf{KD.} KD \cite{kd} aims to reduce the size of a larger model, referred to as the teacher, by transferring its knowledge to a smaller, faster, and similarly performing model known as the student. The fundamental principle behind this approach revolves around compressing the knowledge contained within the teacher model, with the student learning to mimic his predictions \cite{modelcompr2}.
%the predictions made by the teacher.

The temperature parameter $T$ is critical in the knowledge distillation process. It serves as a mechanism to control the level of confidence in the predictions made by the teacher model. A higher temperature leads to a more uniform probability distribution, allowing the student to explore diverse options, while a lower temperature accentuates the differences between classes, focusing on the information deemed more relevant by the teacher (i.e., exploitation) \cite{kd,caragea2}. The temperature adjustment is applied at the softmax function, which computes the probability distribution over classes. Given the input $x_i$, the probability $p_{i,k}$ for class $k$ is computed based on the network logits $z_i$ as follows \cite{kd}:
\begin{equation}
\label{eq:softmax}
p_{i,k} = \frac{\exp (z_{i,k} / T) }{ \sum_{j} \exp (z_{i,j} / T)}
\end{equation}

% TODO: Maybe delete this `dark knowledge` paragraph
% The concept of \textit{dark knowledge}, introduced in \cite{kd}, pertains to the low-probability predictions assigned by the teacher model to incorrect answers. Despite the minimal probabilities associated with these incorrect predictions, they still possess valuable information that can be transmitted to the student model. To illustrate this concept, let us consider an example where an image of a BMW is presented. While it is highly improbable for the teacher model to predict it as a garbage truck, it is even more unlikely for it to mistakenly identify it as a carrot. These low-probability predictions provide valuable insights into the teacher model's reasoning and highlight the distinctions between different classes, thereby enriching the knowledge transfer process during knowledge distillation

The knowledge distillation architecture involves training the teacher and the student neural networks on the same dataset. A hyperparameter $\alpha$ controls the interpolation of partial losses, considering the teacher's soft predictions and the ground truth labels. The distillation loss is calculated using the cross-entropy loss ($\mathcal{L}_{CE}$) between the ground truth and the hard predictions and the Kullback-Leibler (KL) divergence loss ($\mathcal{L}_{KL-KD}$) between the soft labels and soft predictions \cite{caragea2,tannealing}:
\begin{equation}
    \mathcal{L}_{KL-KD} = \frac{T^2}{N} \sum_{i=1}^N KL( p^{t}(x_i, T) \vert\vert p^{s} (x_i, T))
\end{equation}
\begin{equation}
    \mathcal{L}_{KD} = \alpha \mathcal{L}_{CE} + (1 - \alpha) \mathcal{L}_{KL-KD}
\end{equation}
where $p^{t}(x_i, T)$ represents the softmax outputs of the teacher model, and $p^{s}(x_i, T)$ represents the student's softmax output, both computed using Eq. \ref{eq:softmax}.

\textbf{MTKD.} In natural language processing, the simultaneous learning of multiple tasks presents a considerable challenge. MTL addresses this challenge by training a single model to solve numerous tasks concurrently. However, optimizing a model for various tasks with different complexities can result in performance imbalances, where specific tasks dominate while others suffer \cite{mtkd}.

To tackle the performance imbalance problem in MTL scenarios, multi-task learning with knowledge distillation (MTKD) has been proposed by \cite{mtkd_born}. This approach leverages the benefits of both techniques to overcome the performance imbalance problem in MTL scenarios. The core idea is to use specialized models, called single-task teacher models, to teach a multi-task student model. The teacher models provide rich information beyond simple one-hot encodings, and this knowledge is transferred to the student model through distillation.

Based on the findings in \cite{caragea}, let $\mathcal{D}^\tau = {\{ (x_{i}^{\tau}, y_{i}^{\tau}) \}}$ with $N^\tau$ examples represent the training dataset for the  task $\tau$, $\theta$ represents student's parameters being updated, and $\theta^{\tau}$ represents teacher's parameters. We denote $f^{\tau}(x_{i}^{\tau}, \theta^{\tau})$ the output computed using Eq. \ref{eq:softmax} of each task-specific teacher model specialized in task $\tau$, trained using fine-tuned BERT, and $f^{\tau}(x_{i}^{\tau}, \theta)$ the output according to Eq. \ref{eq:softmax} of the student model on task $\tau$. The loss is described as follows \cite{caragea,caragea2,tannealing}:
\begin{equation}
\mathcal{L}_{KL-MTKD}(\theta) = \sum_{\tau = 1}^{4} \frac{T^2}{N^{\tau}} \sum_{(x_{i}^{\tau}, y_{i}^{\tau}) \in \mathcal{D}^\tau} KL (f^{\tau}(x_{i}^{\tau}; \theta^{\tau}) \vert\vert f^{\tau}(x_{i}^{\tau}; \theta))
\end{equation}
\begin{equation}
\mathcal{L}_{MTKD} = \alpha \mathcal{L}_{CE} + (1 - \alpha) \mathcal{L}_{KL-MTKD}
\end{equation}

\begin{figure*}[!t]
    \centering
    \includegraphics[width=\textwidth]{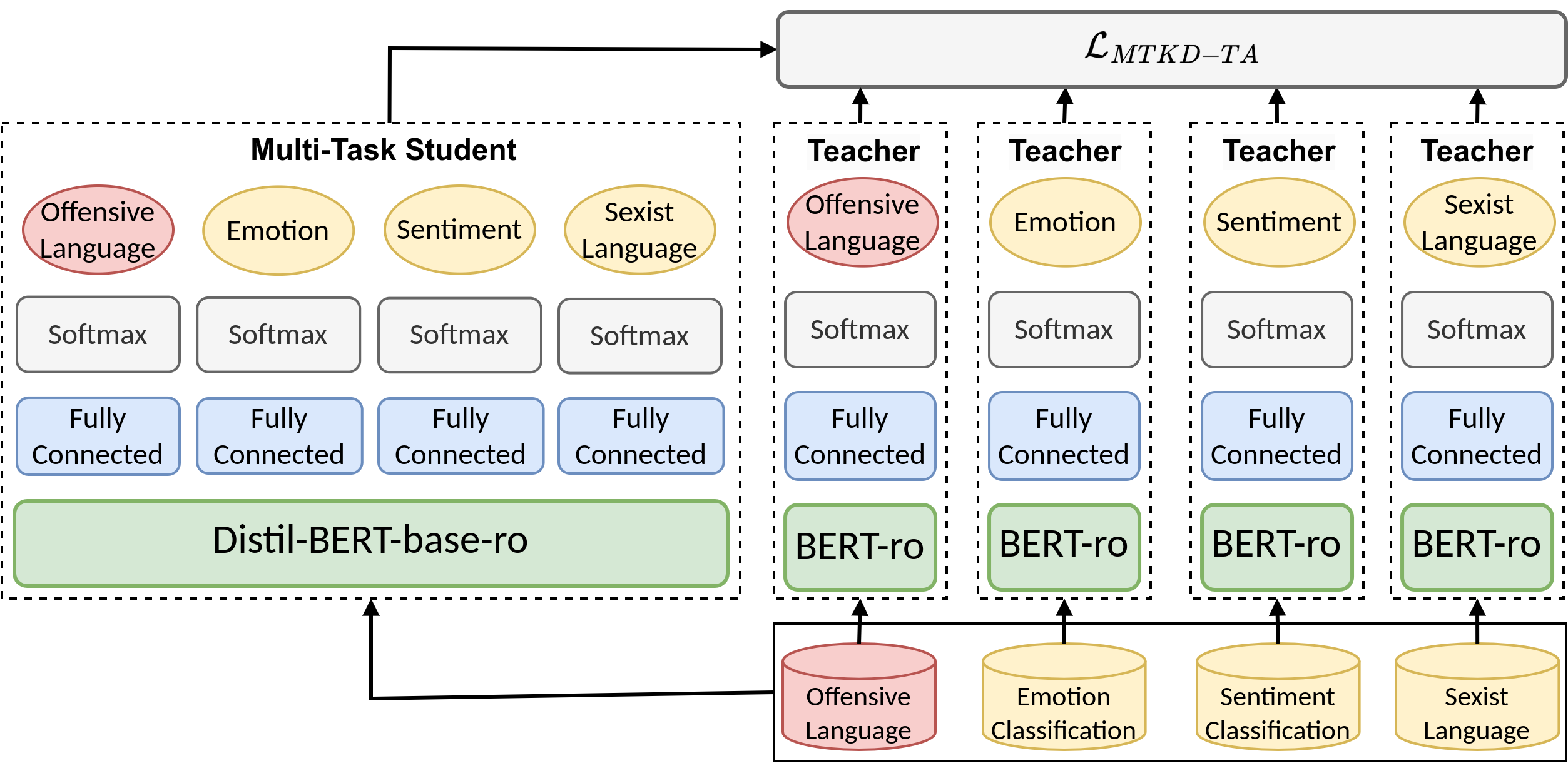}
    \caption{Multi-task learning with knowledge distillation and teacher annealing.}
    \label{kd-mtl}
\end{figure*}

\textbf{MTKD-TA.} Teacher annealing (TA) \cite{tannealing} is an optimization technique employed in conjunction with the knowledge distillation method to handle better the discrepancies between the student and the teacher models. While temperature is typically used to alleviate this issue, \cite{tannealing_asistent} highlights that as the capacity of the teacher model increases, thereby accentuating the differences with the student model, the student's performance improves only up to a certain point, after which it decreases.

The TA method addresses the capacity difference problem in knowledge distillation by gradually reducing the influence of the teacher model. In contrast to the teacher-assistant knowledge distillation approach, which introduces an intermediate network, teacher annealing relies on increasing linearly a parameter during training \cite{caragea,mtkd_born}, called $\lambda$, from 0 to 1. It controls the balance between the distillation loss and the supervised loss, which measures the discrepancies between the student's predictions and the teacher's predictions and between the student's predictions and the ground truth labels, respectively \cite{tannealing}. By gradually decreasing the teacher's influence and increasing reliance on the original labels, the student model becomes more independent and capable of achieving improved performance  \cite{mtkd_born}. Thus, we modify the multi-task learning with knowledge distillation and teacher annealing (MTKD-TA) loss function as follows:
\begin{equation}
\mathcal{L}_{MTKD-TA} = \lambda \mathcal{L}_{CE} + (1 - \lambda) \mathcal{L}_{KL-MTKD}
\end{equation}

As depicted in Fig. \ref{kd-mtl}, the final architecture incorporates four task-specific datasets to obtain individual teacher models by fine-tuning BERT. Each teacher model is specialized for one of the four tasks. The student model adopts an MTL architecture with a modified weight updating scheme in its network, accounting for the newly introduced loss calculation formula.

\section{Experiments}
\subsection{Datasets}
%The main focus of this research was to develop a model for automatic detection of Romanian offensive language and its subtypes. 
The target dataset used in this research was the RO-Offense dataset\footnote{\url{https://huggingface.co/datasets/readerbench/ro-offense}}, consisting of relevant examples curated explicitly for the proposed task. Additionally, three auxiliary datasets were included to enhance the model's performance in achieving the main objective. By incorporating these additional datasets, the aim is to improve the accuracy and generalization capability of the model in effectively identifying and classifying offensive language.

\textbf{RO-Offense.} RO-Offense is the largest publicly available dataset for analyzing offensive discourse in the Romanian language. It comprises 12,447 annotated records, classified into four distinct classes: \textit{Profanity} (13\%), \textit{Insult} (23\%), \textit{Abuse} (28\%), and \textit{Other} (36\%). To ensure privacy, the dataset has anonymized names of individuals and organizations, replacing them with generic labels.

\textbf{REDv2.} The Romanian Emotions Dataset (REDv2) \cite{redv2} is a publicly available dataset, hosted on GitHub\footnote{\url{https://github.com/Alegzandra/RED-Romanian-Emotions-Dataset}}, that provides 5,449 manually verified tweets for analyzing emotions in the Romanian language. Each example in the dataset is classified into one of seven possible classes: \textit{Anger}, \textit{Fear}, \textit{Joy}, \textit{Sadness}, \textit{Surprise}, \textit{Trust}, and \textit{Neutral}. Additionally, all tweets have been anonymized by removing usernames and proper nouns from the dataset.

\textbf{CoRoSeOf.}\quad\quad The Corpus of Romanian Sexist and Offensive language (CoRoSeOf) \cite{coroseof} is a publicly available dataset that is a valuable resource for studying sexist and offensive language in the Romanian context. The dataset, which can be found on GitHub\footnote{\url{https://github.com/DianaHoefels/CoRoSeOf}}, contains 39,245 tweets with labels assigned by multiple annotators for the classification of sexist and offensive language in Romanian. Initially, each instance in the dataset was assigned to one of the five possible classes: \textit{Direct Sexism}, \textit{Descriptive Sexism}, \textit{Reportive Sexism}, \textit{Non-Sexist Offensive}, and \textit{Non-Sexist}. However, for this research, the data has been transformed into a binary classification format, where all sexist subtypes are included in the \textit{sexist} class, while the remaining instances are included in the \textit{non-sexist} class.

\textbf{LaRoSeDa.} The Large Romanian Sentiment Data Set (LaRoSeDa) \cite{laroseda} is a publicly available resource, accessible on GitHub\footnote{\url{https://github.com/ancatache/LaRoSeDa}}, that consists of 15,000 reviews collected from one of the largest e-commerce platforms in Romania. Each instance in the dataset is labeled as either \textit{positive} or \textit{negative}, allowing for sentiment analysis and contributing to a better understanding of the sentiment patterns in the Romanian language.

\subsection{Experimental Settings}
% The Ro-Offense dataset was split into 80\% for training, 10\% for validation, and 10\% for testing. 
The REDv2 dataset was split into 75\% for training, 10\% for validation, and 15\% for testing. For Ro-Offense, CoRoSeOf, and LaRoSeDa datasets, we use the 80\%/10\%/10\% split. All the trained models used the Transformer library\footnote{\url{https://github.com/huggingface/transformers}} as their base architecture, and their versions are managed using the HuggingFace platform\footnote{\url{https://huggingface.co/}}. 
The base architecture used for the distilled student model is Distil-BERT-base-ro\footnote{\url{https://huggingface.co/racai/distilbert-base-romanian-cased}}, while the other models utilize BERT-base-ro-cased\footnote{\url{https://huggingface.co/dumitrescustefan/bert-base-romanian-cased-v1}} (BERT-ro). The main difference between these two architectures is the number of layers \cite{arhitectura_bert}. Distil-BERT-base-ro consists of 6 layers, 81M parameters, and requires 312MB of memory, whereas BERT-ro comprises 12 layers, 124M parameters, and occupies 477MB of memory.
The configuration includes a starting learning rate of 2e-5, AdamW optimizer, weight decay of 0.01, and batch size 16. The number of fine-tuning epochs varies between 2 and 7, depending on the dataset size and the architecture on which the model was trained. The probability $\alpha$ used in noisy student takes values from the set \{15, 20, 25\}. The interpolation parameter $\lambda$ used in MixUp is set to either 15 or 30.
For model evaluation, we use accuracy (Acc), precision (P), recall (R), and weighted $F_1$-score ($F_1$).

\section{Results}

\subsection{Results for Knowledge Distillation Models}

We focus on the distillation technique combined with multi-task learning, which enables the transfer of information into a model of the same size but benefits from diverse inputs from multiple sources of knowledge. This approach allows for the development of a compact model that can maintain the high performance of its teacher. The results are presented in Table \ref{kd-mtl-ablation}. For the multi-task learning experiments, we employ the REDv2, LaRoSeDa, and CoRoSeOf datasets as auxiliary tasks.

\textbf{Fine-Tuning BERT.} During experiments, we noticed that the offensive language detection model based on the BERT-ro architecture, specifically trained for offensive language detection and its subtypes, achieves commendable results with an accuracy of 78.63\% and an $F_1$-score of 78.83\%. However, BERT-ro is not easily saturable, and there is room for further enhancement. These scores indicate the potential for optimizing the model to achieve even better results.

\begin{table*}[!b]
\caption{Results on the RO-Offense dataset for knowledge distillation approaches.}
\label{kd-mtl-ablation}
\centering
\setlength{\tabcolsep}{5pt}
\begin{tabular}{lcccc}
\thickhline
Model & Acc & $F_1$ & P & R  \\ \thickhline
\multicolumn{5}{c}{\light{\textit{base model}} } \\
BERT-ro & 78.63 & 78.83 & 79.15 & 78.63 \\
MTL & 77.99 & 77.85 & 77.80 & 77.99 \\ \hline
\multicolumn{5}{c}{\light{\textit{distilled student}} } \\
KD & 77.10 & 77.23 & 77.44 & 77.10 \\
MTKD & 81.36 & 81.19 & 81.12 & 81.36 \\
MTKD-TA & \textbf{82.40} & \textbf{82.34} & \textbf{82.29} & \textbf{82.40} \\
\thickhline
\end{tabular}
\end{table*}

\textbf{KD.} The results obtained by the student using the KD technique on the main dataset are lower than those achieved by the BERT-ro model. There is an approximate 1.5\% drop in all evaluated metrics, which can be intuitively explained by the fact that although the student benefits from both the soft probabilities from the teacher and the direct information from the dataset, it fails to reach the same performance due to the reduced size of the architecture of Distil-BERT-base-ro, which consists of only 6 layers instead of 12.

\textbf{MTL.} The MTL model combines all datasets and relies on the larger architecture, BERT-ro. According to Table \ref{kd-mtl-ablation}, the results obtained by the MTL model are superior to those of the KD model, as the larger architecture allows for more complex learning. However, the MTL model still falls short of the performance achieved by the teacher model. This can be attributed to the inherent challenge of simultaneously learning multiple tasks, mainly when dealing with larger datasets. Managing each task's contribution and finding a learning balance is crucial to improving overall performance.

\textbf{MTKD.} The MTKD model significantly improves over prior experiments. It surpasses the performance of the teacher model by $\sim$2.3\%, the distilled student by $\sim$3.9\%, and the MTL model by $\sim$3.3\%. This improvement is achieved by leveraging the transfer of knowledge from the teacher through a processing step at a temperature $T=4$ and utilizing the ground truth information. The balance between these two sources of information is achieved through $\alpha=0.6$ controlling the partial interpolation of losses.

\textbf{MTKD-TA.} The MTKD-TA model showcases two significant aspects based on the results obtained. First, as we showed in the previous experiments, the difference in architecture size led to a loss of information transferred from the teacher to the student, which was partially regained in this experiment. Second, we can achieve better results by dynamically scaling the $\lambda$ coefficient. For most tasks, the coefficient $\lambda$ is increased incrementally between 0 and 1, with the temperature fixed at $T=2$. For the emotion detection task, the temperature $T=7$ is more effective. The MTKD-TA model outperforms MTKD by approximately 1\%, the distilled student by around 5\%, and BERT-ro by roughly 3.5\%.

\subsection{Impact of Data Augmentation}

We first explore distilling the base model into a smaller and faster model, which also benefits from various data augmentation techniques to enhance its performance. By employing these techniques, the aim is to strike a balance between efficiency and accuracy. We present the results of applying data augmentation techniques to the model obtained through distillation on the smaller architecture, Distil-BERT-base-ro. The results are summarized in Table \ref{dataaugtable}.

\begin{table*}[!t]
\caption{Results of student data augmentations on the RO-Offense dataset.}
\label{dataaugtable}
\centering
\begin{tabular}{lcccc}
\thickhline
Model & Acc & $F_1$ & P & R  \\ \thickhline
KD & 77.10 & 77.23 & 77.44 & 77.10 \\
\quad \textit{+MixUp Encoder} & 77.02 & 77.21 & 77.53 & 77.02 \\
\quad \textit{+MixUp Sent.-30\%} & 77.51 & 77.63 & 77.83 & 77.51 \\
\quad \textit{+RoGPT-2} & 77.51 & 77.71 & 78.10 & 77.51 \\
\quad \textit{+ASDA} & 77.75 & 77.73 & 77.81 & 77.75 \\
\quad \textit{+Noisy-25\%} & 78.07 & 78.08 & 78.18 & 78.07 \\
\quad \textit{+ASDA+RoGPT-2} & 77.67 & 77.81 & 78.14 & 77.67 \\
\quad \textit{+ASDA+RoGPT-2+Noisy-15\%} & 77.91 & 78.11 & 78.54 & 77.91 \\
\quad \textit{+ASDA+RoGPT-2+Noisy-20\%} & 78.55 & 78.68 & 78.90 & 78.55 \\
\quad \textit{+ASDA+RoGPT-2+Noisy-20\%+MixUp Sent.-30\%} & 78.07 & 78.30 & 78.74 & 78.07 \\
\quad \textit{+ASDA+RoGPT-2+Noisy-20\%+MixUp Sent.-15\%} & \textbf{78.71} & \textbf{78.81} & \textbf{79.06} & \textbf{78.71} \\
\thickhline
\end{tabular}
\end{table*}

\textbf{MixUp Encoder.} This method does not show better results in our results. The evaluation metrics do not exhibit significant differences that warrant considering this method in combination with other data augmentation techniques.

\textbf{MixUp Sentence.} Applying the MixUp Sentence technique with interpolation 30\% (i.e., MixUp Sent.-30\%) at a higher level led to an accuracy improvement of approximately 0.41\% compared to the distilled student model alone. This significant difference justifies combining this method with other augmentation techniques.

\textbf{RoGPT-2.} We employ RoGPT-2 to replace 30\% of the end of the texts. This augmentation technique results in a performance improvement of approximately 0.5\% compared to the reference model, similar to the MixUp Sentence technique. The metrics indicate the potential benefits of combining it with other augmentation methods. However, although more diverse, the augmentation is riskier, completing the sentences more creatively.

\textbf{ASDA.} The utilization of the ASDA method results in an improvement of at least 0.5\% in evaluation metrics. This approach is considered safe and provides a richer learning context.

\textbf{Noisy Student.} As observed in Table \ref{dataaugtable}, the noisy student (i.e., Noisy) augmentation technique proves to be the most effective. This method achieves a significant increase in results of at least 0.8\% compared to the distilled student alone. Despite its simplicity, the method's performance underscores the significance of obtaining controlled noise. In this case, the constraint involves introducing a 25\% probability of change, both for word elimination and potential sentence completion.

\textbf{Combining augmentation techniques.} After analyzing each augmentation technique individually, we combined ASDA and RoGPT-2 to balance context and creativity, resulting in a 0.6\% increase over the student model. Then, noise was added with a 15\% probability initially, but a 20\% probability yielded a 1.4\% increase. Ultimately, the best-performing approach combines ASDA, RoGPT-2, Noisy Student, and MixUp Sentence. Regarding the MixUp Sentence method, much better outcomes are obtained by reducing the probability of interpolating examples from 30\% to 15\%. In the end, we achieve an advantage improvement of 1.58\% over the distilled student alone, equalizing the performance of the BERT-ro teacher, which has 53\% more parameters.

% Based on the individual analysis of each augmentation technique, it was deemed appropriate to combine the \textit{ASDA} and \textit{RoGPT-2} methods to achieve a balance between a method that provides more context and one that is more creative. The results highlighted a significant \textit{increase} of approximately 0.6\% compared to the distilled student alone, partially benefiting from the advantages brought by each individual method.

% Subsequently, the combination was performed together with the Noise technique, initially choosing an augmentation probability of 15\% to avoid excessive model noise. However, the experiment demonstrated that a significant performance increase of approximately 1.4\% compared to the reference can be achieved by using a 20\% augmentation probability. 

\subsection{Impact of Auxiliary Tasks}

Through the lens of the ablation study, we analyze the behavior of MTL models on offensive language detection by removing different combinations of tasks. The results presented in Table \ref{kd-mtl-ablation2} pertain to the evaluation using the $F_1$-score. The \textit{proposed model} refers to the models that employ all three auxiliary tasks, namely emotions detection, sentiment classification, and sexist language detection. Then, we present the results obtained after removing the specified tasks, enumerated after \textit{w/o} (i.e., without). Note that the \textit{proposed model} without any auxiliary tasks in the MTL setting is equivalent to the BERT-ro model.

\textbf{Sexist language.} We notice a significant variation in the impact of the sexist language task on the final outcome. In the case of the MTL model, the exclusion of this task leads to very poor results, while for the MTKD-TA model, the results are quite good even without considering this task. One possible explanation could be the difficulty of accommodating a larger dataset within the MTL environment.

\textbf{Combining dataset exclusions.}
We notice the combinations $\{$emotions, sexist language$\}$ and $\{$sentiment, sexist language$\}$ yield similar results, indicating that the model can successfully learn even without one of the tasks that analyze emotions and sentiments. Additionally, a notably lower performance is observed for the MTKD-TA model when both tasks are excluded from the analysis.

\begin{table*}[!t]
\caption{Results on RO-Offense after removing auxiliary tasks.}
\label{kd-mtl-ablation2}
\centering
\begin{tabular}{lccc}
\thickhline
Model & MTL & MTKD & MTKD-TA \\ \thickhline
\textit{Proposed model}  & 77.85 & 81.19 & 82.34 \\
        {\quad \textit{w/o} emotions \& sentiment \& sexist language}    & 78.83 & 77.23 & - \\ 
        {\quad \textit{w/o} emotions \& sentiment}              & 80.08 & 81.35 & 80.69 \\
        {\quad \textit{w/o} emotions \& sexist language}              & 78.97 & \textbf{81.74} & 82.26 \\
        {\quad \textit{w/o} sentiment \& sexist language}           & 78.25 & 81.67 & \textbf{82.39} \\ 
        {\quad \textit{w/o} emotions}                        & \textbf{80.82} & 81.47 & 81.53 \\
        {\quad \textit{w/o} sentiment}                     & 79.17 & 81.14 & 81.69 \\ 
        {\quad \textit{w/o} sexist language}                     & 78.71 & 80.97 & 81.97 \\
\thickhline
\end{tabular}
\end{table*}

% \section{Limitations}
% Experiments and resulting models have notable limitations.

% Firstly, despite diligent efforts, we encountered \textbf{challenges in identifying a data augmentation method} that effectively enhances the performance of the distilled models trained using the MTL and TA techniques. This difficulty may stem from the intricate task of striking an \textbf{optimal balance} in harnessing the contributions of individual datasets to achieve comprehensive  generalization for the main task.

% Secondly, it is crucial to acknowledge the potential existence of \textbf{superior combinations of hyperparameters} that could yield improved results. However, the exploration of hyperparameters was constrained by the limited availability of hardware resources.

\section{Conclusion}
This paper developed neural network models to detect Romanian offensive language and investigated various techniques to enhance their performance. Integrating additional related tasks (i.e., emotion analysis, sentiment analysis, and sexist language detection) through MTL demonstrated improved performance. However, achieving an optimal balance between the contributions of different tasks in the MTL environment proved challenging. To address this, we employed KD and TA, resulting in $\sim$3.5\% performance improvement compared to the BERT-ro model. Additionally, efforts were made to reduce the model size and utilize data augmentation techniques, leading to an additional performance increase of $\sim$1.6\%.

Future research directions involve exploring more diverse datasets, optimizing the MTL setup, fine-tuning hyperparameter combinations, and considering alternative base architectures. These advancements aim to strengthen the detection and effective management of Romanian offensive language, contributing to content filtering, and establishing a safer virtual environment.

\section*{Acknowledgements}
This work was supported by the NUST POLITEHNICA Bucharest through the PubArt program, and a grant from the National Program for Research of the National Association of Technical Universities - GNAC ARUT 2023.

\bibliographystyle{splncs04}
\bibliography{biobio.bib}

\end{document}